\documentclass[conference,a4paper,10pt]{IEEEtran}
\IEEEoverridecommandlockouts
\usepackage{cite}
\usepackage{amsmath,amssymb,amsfonts}
\usepackage{graphicx}
\usepackage{textcomp}
\usepackage{xcolor}
\def\BibTeX{{\rm B\kern-.05em{\sc i\kern-.025em b}\kern-.08em
    T\kern-.1667em\lower.7ex\hbox{E}\kern-.125emX}}
\usepackage{balance}
\usepackage{comment}
\usepackage[tight,footnotesize]{subfigure}
\usepackage[active]{srcltx}
\usepackage{mathtools}

\graphicspath{{./figs/}}
\usepackage{eurosym}
\usepackage[plain]{algorithm}
\usepackage{multicol}
\usepackage{dsfont}
\usepackage{cases}
\usepackage{algpseudocode} 
    
\begin{document}

\title{Synergising Human-like Responses and Machine Intelligence for Planning in Disaster Response\\
\thanks{This work is supported by the European Union's Horizon 2020 research and innovation programme under grant agreement No 739551 (KIOS CoE), and the Republic of Cyprus through the Deputy Ministry of Research, Innovation and Digital Policy.}
}

\author{
	\IEEEauthorblockN{
		Savvas~Papaioannou,
		Panayiotis~Kolios,
        Christos~G.~Panayiotou, and
		Marios~M.~Polycarpou
	}
	\IEEEauthorblockA{
		\textit{KIOS Research and Innovation Center of Excellence}\\
		\textit{Department of Electrical and Computer Engineering}\\
        University of Cyprus, Nicosia, Cyprus\\
		\{papaioannou.savvas, pkolios, christosp, mpolycar\}@ucy.ac.cy
	}
}


\maketitle

\begin{abstract}
In the rapidly changing environments of disaster response, planning and decision-making for autonomous agents involve complex and interdependent choices. Although recent advancements have improved traditional artificial intelligence (AI) approaches, they often struggle in such settings, particularly when applied to agents operating outside their well-defined training parameters. To address these challenges, we propose an attention-based cognitive architecture inspired by Dual Process Theory (DPT). This framework integrates, in an online fashion, rapid yet heuristic (human-like) responses (System 1) with the slow but optimized planning capabilities of machine intelligence (System 2). We illustrate how a supervisory controller can dynamically determine in real-time the engagement of either system to optimize mission objectives by assessing their performance across a number of distinct attributes. Evaluated for trajectory planning in dynamic environments, our framework demonstrates that this synergistic integration effectively manages complex tasks by optimizing multiple mission objectives.
\end{abstract}


\section{Introduction} \label{sec:Intro}

As of today, the integration of technology and automation into disaster response missions is in its early stages \cite{Kyrkou2023}.  For instance, despite the proposal of autonomous agents, such as unmanned aerial vehicles (UAVs), for a wide range of applications \cite{PapaioannouIROS2021,PapaioannouTAES,PapaioannouICUAS2022}, their use in disaster response scenarios remains limited. A critical function of autonomous agents in disaster response is devising plans to reach a goal region or explore unknown, cluttered environments, often while searching for victims \cite{Liu2013,PapaioannouTSMC}. The task of trajectory planning \cite{Madridano2021}, is a crucial component for robotic applications and automation in general. It is now a major challenge in artificial intelligence (AI), especially in disaster response scenarios where complexity stems from the unpredictable and dynamic nature of the environments and the limited decision-making capabilities of the agents in planning and control. 

\begin{figure}
	\centering
	\includegraphics[scale=0.4]{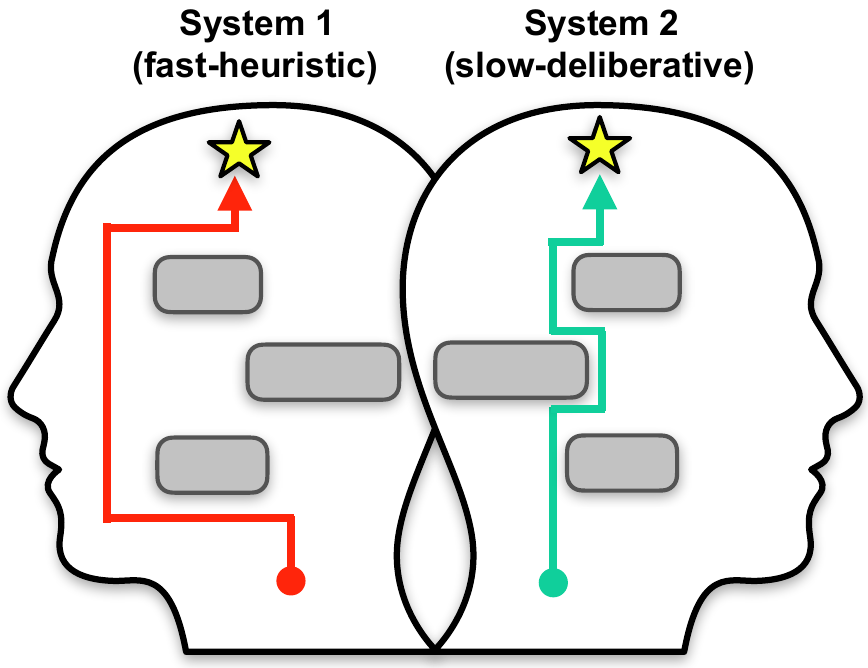}
	\caption{Rapid responses (System 1), and rational reasoning (System 2) in human cognition.}	
	\label{fig:fig1}
	\vspace{-3mm}
\end{figure}

Drawing inspiration from human attributes such as intuition, rapid response, and adaptability, our key observation is that many challenging tasks, particularly in disaster response planning, can be effectively tackled by combining human-like reasoning with machine intelligence. This concept is rooted in the Dual Process Theory (DPT) \cite{Evans2009}, which posits two distinct modes of thinking in human cognition (Fig. \ref{fig:fig1}): the intuitive, heuristic, and fast System 1, and the more deliberate, analytical, and slow System 2. Designing intelligent machines by studying the human mind and drawing inspiration from human cognitive processes is a foundational concept that intersects multiple disciplinary areas \cite{Butlin2023,Sloman1999,Velik2011}. 

Related to our work is the study by Lin et al. \cite{Lin2023}, which, drawing on the concepts of fast and slow thinking, introduced a generative AI agent for planning actions in interactive reasoning tasks. In a similar manner, Anthony et al. \cite{Anthony2017} developed a DPT-inspired reinforcement learning (RL) algorithm. This algorithm utilizes two distinct action selection mechanisms: System 1, which operates quickly without lookahead, and System 2, which employs a more deliberate lookahead approach via Tree-Search. Furthermore, the research by Booch et al. \cite{Booch2021} builds on cognitive theories of human decision-making, exploring various methods to enhance artificial intelligence. Moreover, in \cite{Gulati2021} the authors investigate different rule-based switching mechanisms for transitioning control between the fast System 1 and the slow System 2 in the game of Pac-Man, whereas in \cite{Dalmasso2021} path-planning for a robot is managed by the autonomous system (i.e., System 1) until a collision is detected, at which point a human agent (i.e., System 2) takes control of the planning process. 

Complementary to the works mentioned above, our approach aims to synergize rapid human-like decision-making with the processing capabilities of machine intelligence. Specifically, we propose a cognitive architecture that integrates rapid human-like responses with analytical problem-solving capabilities of machine intelligence. We present a decision-making (switching between System 1 and System 2) and planning (trajectory optimization) framework based on the dual-process theory, tackled by a cognitive controller. We design an attention-based supervisory controller to oversee the interplay between System 1 and System 2, and finally, we evaluate the proposed framework for trajectory planning in dynamic environments, demonstrating the synergistic effect of integrating the complementary capabilities of the two systems.

The rest of the paper is organized as follows: Section \ref{sec:problem} develops the system model and formulates the problem tackled. Section \ref{sec:approach} discusses the details of the proposed approach, Section \ref{sec:Evaluation} evaluates the proposed approach, and finally, Section \ref{sec:conclusion} concludes the paper.

\section{Problem Statement} \label{sec:problem}

In this study, we address a challenge inspired by real-world emergency response scenarios, specifically during wildfire disasters. The task involves guiding a UAV agent with shared-autonomy capabilities to search for survivors in remote, wildfire-affected forest areas. The UAV's objective is to reach a predetermined goal region, ensuring the avoidance of firefronts along its path.
We assume that a disaster early-warning system (EWS) is in place, equipped with a variety of sensors and data sources, including weather stations and satellite imagery. This system provides real-time alerts and forecasts the propagation of firefronts, thereby equipping the rescue team with comprehensive environmental data. Consequently, the team's task is to devise a UAV trajectory that not only reaches the target area but also navigates safely through the dynamic disaster environment.
The UAV is designed to operate in two distinct modes: semi-autonomous (System 1) and autonomous (System 2). In the semi-autonomous mode, first responders input waypoints into a mobile device. The UAV then follows these waypoints, thereby guiding the agent to the destination. This approach, denoted as System 1, allows for rapid response but lacks optimization for mission completion time and does not account for the UAV's battery life and energy consumption. Conversely, the autonomous mode (System 2) employs an optimal controller that, considering both UAV dynamics and firefront conditions, computes the UAV's control inputs over a planning horizon optimizing either the mission completion time or the UAV's energy consumption. 

\begin{figure}
	\centering
	\includegraphics[width=\columnwidth]{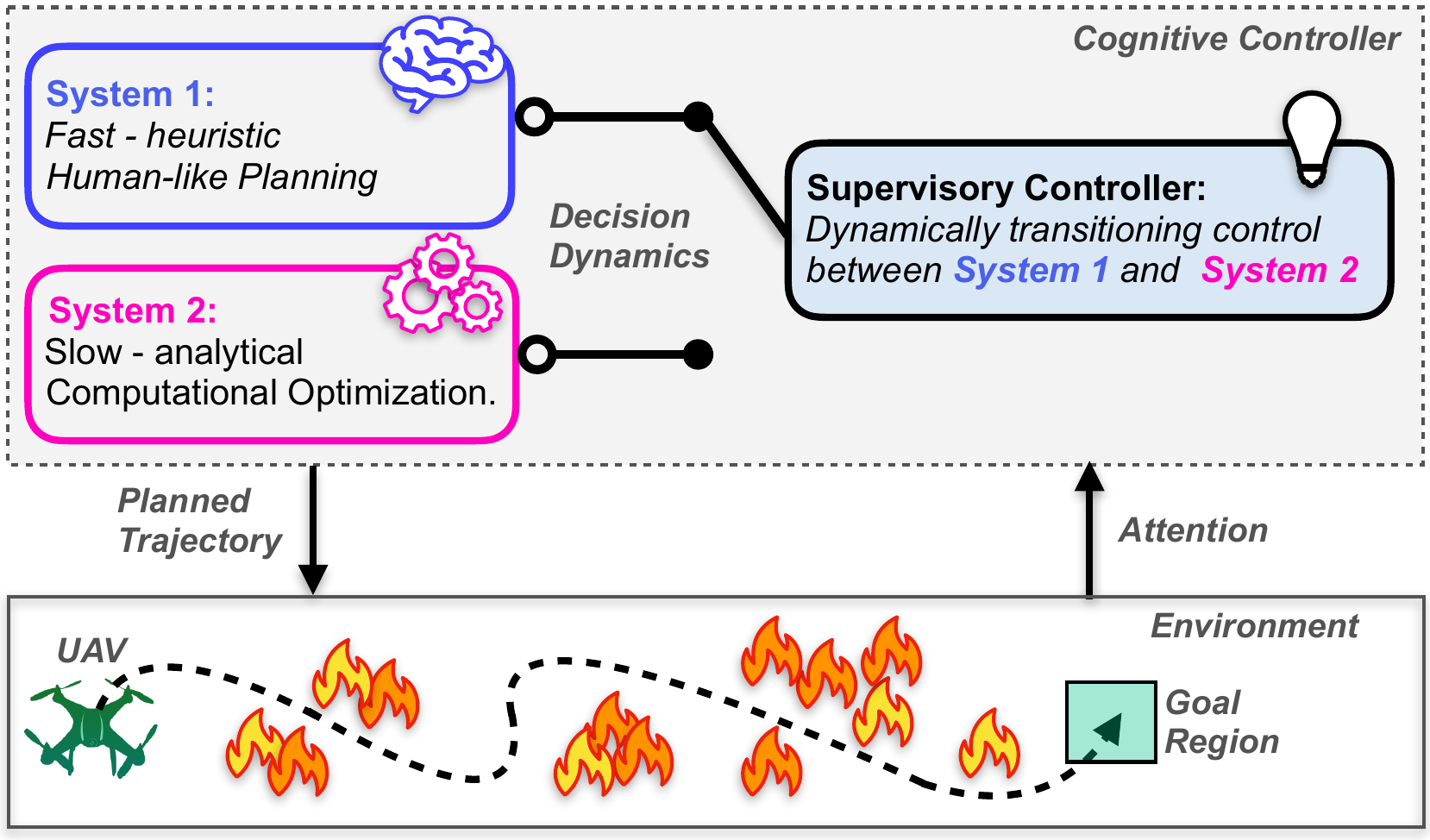}
	\caption{The figure illustrates the proposed tripartite cognitive architecture.}	
	\label{fig:fig2}
	\vspace{-5mm}
\end{figure}

In this work, we have developed both System 1 and System 2, and subsequently, we have devised a supervisory controller. This controller employs an attention mechanism to dynamically switch control between the two systems in response to changing environmental conditions. This tripartite system architecture (i.e., termed hereafter as cognitive controller), is depicted in Fig. \ref{fig:fig2}.

\subsection{Agent Dynamics} \label{ssec:agent_dynamics}
We consider that a UAV agent, denoted as $x_{t}$, is operating within a bounded 3D stochastic disaster environment $\mathcal{E} \subset \mathbb{R}^3$. This agent follows discrete-time dynamics, represented by a linear state-space model:
\begin{equation}\label{eq:kinematics}
x_{t} = 
\begin{bmatrix}
    I_{3\times3} & \Delta T \cdot I_{3\times3}\\
    0_{3\times3} & (1-\zeta) \cdot I_{3\times3}
\end{bmatrix} x_{t-1} + 
\begin{bmatrix}
    0_{3\times3} \\
    \frac{\Delta T}{m} \cdot I_{3\times3}
\end{bmatrix} u_{t},
\end{equation}

\noindent where the system evolution is concisely described as $x_{t}=\Phi x_{t-1}+\Gamma u_{t}$. Here, $x_t = [(x_t^{\mathbf{p}})^\top, (x_t^{\mathbf{v}})^\top]^\top \in \mathcal{X} \subset \mathbb{R}^6$ represents the state of the agent at time-step $t$. This state comprises its position ($x_t^{\mathbf{p}} \in \mathbb{R}^3$) and velocity ($x_t^{\mathbf{v}} \in \mathbb{R}^3$) components within the 3D Cartesian coordinate system. The agent is assumed to be controllable, capable of following specific directional and speed commands via the control input $u_t \in \mathcal{U} \subset \mathbb{R}^3$, which corresponds to the applied force. In Eq. \eqref{eq:kinematics}, $\Delta T$ signifies the sampling interval, $\zeta$ is the air resistance coefficient, and $m$ represents the mass of the agent. Additionally, $I_{3\times3}$ and $0_{3\times3}$ are the 3-by-3 identity and zero matrices, respectively.

\subsection{Firefront Dynamics} \label{ssec:firefront_dynamics}
In our model, we consider that at each time step $t$, a stochastic number $N_t$ of firefronts $Y^i_t, i \in\{1,..,N_t\}$ emerge and propagate within the disaster environment $\mathcal{E}$. The genesis and dissipation of these firefronts are modelled as stochastic processes, influenced by a set of variable environmental factors present at each time step. These factors include, but are not limited to, the intensity and direction of prevailing winds, the availability and type of combustible materials (fuel), and ambient temperature and humidity conditions. The propagation dynamics of each firefront are governed by a probabilistic dynamical model, which integrates these environmental variables to predict the movement and intensity of the firefronts over time. This model accounts for the inherent unpredictability and spatial-temporal variability of fire behavior in disaster environments. 

Specifically, the state of a firefront at time $t$, denoted as $Y_t$ (where we have dropped index $i$ for notational clarity), is modelled as a Bernoulli random process \cite{Ristic2013}, with dynamics governed by the transitional density $\omega_{t|t-1}(Y_t|Y_{t-1})$ given by:
%
\begin{center}
\vspace{+0mm}
\begin{tabular}{ l|c|c| }
\multicolumn{1}{r}{}
 &  \multicolumn{1}{c}{$Y_t =\emptyset$}
 & \multicolumn{1}{c}{$Y_t =\{y_t\}$} \\
\cline{2-3}
$Y_{t-1} =\emptyset$ & $1-p_b$ & $p_b b_t(y_t)$ \\
\cline{2-3}
$Y_{t-1} =\{y_{t-1}\}$& $1-p_s$ & $p_s \pi_{t|t-1}(y_t|y_{t-1})$  \\
\cline{2-3}
\end{tabular}
\vspace{+0mm}   
\end{center}

\noindent The spatial state of the firefront is denoted as $y_t = [y^{\nu}_t, y^{d}_t]^\top$ where $y^{\nu}_t$ and $y^{d}_t$ denote the firefront propagaion speed on the ground, and direction respectively. The term $p_b$ denotes the probability of firefront genesis, $b_t(.)$ is the firefront genesis density, which without loss of generality is assumed to be uniform inside the surveillance area, and $p_s$ is the probability that the firefront propagates to the next time-step. Finally, $\pi_{t|t-1}(y_t|y_{t-1})$ is the firefront propagation density on the ground which in this work it is assumed to be governed by a Gauss-Markov stochastic dynamical model of the form:
\begin{subequations}
\begin{align}
	y^{\nu}_t &= \gamma y^{\nu}_{t-1} + (1-\gamma)\bar{y}^{\nu} + \sqrt{(1-\gamma^2)}\hat{y}^{\nu},\\
	y^{d}_t &= \gamma y^{d}_{t-1} + (1-\gamma)\bar{y}^{d} + \sqrt{(1-\gamma^2)}\hat{y}^{d},
\end{align}
\end{subequations}

\noindent where $\bar{y}^{\nu}$ and $\bar{y}^{d}$ are the mean value of speed and direction respectively, $\hat{y}^{\nu} \sim \mathcal{N}(0,\sigma_\nu)$, and $\hat{y}^{d} \sim \mathcal{N}(0,\sigma_d)$ are Gaussian zero mean random variables with standard deviations $\sigma_\nu$, and $\sigma_d$ respectively, and $\gamma \in [0,1]$ is a design parameter which models randomness. 

Additionally, we assume that at each time-step \( t \), a firefront \( Y_t \) occupies a distinct volume within the environment \( \mathcal{E} \). This volume is mathematically represented as a time-varying rectangular cuboid \( C_t \), characterized by its dynamically changing dimensions. Specifically, a point in three-dimensional space, denoted as \( p \in \mathbb{R}^3 \), is considered to be inside this cuboid if it satisfies a set of linear inequalities: \( \text{dot}(\alpha_l, p) \leq \beta_l \) for \( l \in \{1,..,6\} \), where $\text{dot}(a,b)$ denotes the dot product between vectors $a$ and $b$. Here, \( \alpha_l \in \mathbb{R}^3 \) represents the outward normal vector to the \( l_\text{th} \) face of the cuboid, and \( \beta_l \) is a corresponding scalar constant. These inequalities collectively define the geometric constraints of the cuboid that encapsulate the spatial extent of the firefront at any given time \(t\). Therefore, a particular cuboid $C_t$, can also be characterised by the parameter set $\{\alpha_l,\beta_l\}_{l=1}^6$.

\subsection{Problem Formulation}
The proposed cognitive controller is depicted below in Problem (P1), in a high-level form.

\begin{algorithm}
\vspace{-5mm}
\begin{subequations}
\begin{align} 
&\hspace*{-3mm}\textbf{(P1)}~\texttt{Cognitive Controller} & \notag\\
& \hspace*{-3mm}~~~~\underset{\{u_{t+\tau|t},x_{t+\tau|t},S_t\}_{\tau=1}^{T}}{\min} ~\mathcal{F}_t(X_T,U_T,S_t), &  \hspace*{-10mm} \label{eq:objective_P1} \\
&\hspace*{-3mm}\textbf{subject to: $\tau \in [1,..,T]$} ~  &\nonumber\\
&\hspace*{-3mm} x_{t+\tau|t} = \Phi x_{t+\tau-1|t} + \Gamma u_{t+\tau|t} & \hspace*{-10mm} \label{eq:P1_1}\\
&\hspace*{-3mm} x_{t|t} = x_{t|t-1} & \hspace*{-10mm} \label{eq:P1_2}\\
&\hspace*{-3mm} x_{t+\tau|t}^{\mathbf{p}} \notin \triangle(C^i_{1:t}) & \hspace*{-10mm} \forall i \label{eq:P1_4}\\
&\hspace*{-3mm} \Psi_t = g\left(\Psi_{t-1},A_t\right) & \hspace*{-10mm} \label{eq:P1_5}\\
&\hspace*{-3mm} S_t = \underset{i \in \{1,2\} }{\arg\max} ~\Psi_t(i) & \hspace*{-10mm} \label{eq:P1_6}\\
&\hspace*{-3mm} x_{t+\tau|t} \in \mathcal{X}, ~u_{t+\tau|t}, \in \mathcal{U},~i \in \{1,..,N_t\} & \hspace*{-10mm} \label{eq:P1_7}
\end{align}
\end{subequations}
\vspace{-5mm}
\end{algorithm}

Specifically, $X_T = \{x_{t+\tau|t}\}_{\tau=1}^{T}$, $U_T = \{u_{t+\tau|t}\}_{\tau=1}^{T}$, and $S_t \in \{1, 2\}$ signifies the active system (i.e., System 1 or System 2), chosen by the supervisory controller, for trajectory generation of the agent. The term $x_{t'|t}$ denotes the predicted state of the agent at a future time-step $t'$, computed at the current time-step $t$. Problem (P1) is activated at each time-step $t$ to determine the agent's control inputs $u_{t+\tau|t}$ for a planning horizon of length $T$ time-steps i.e., $\tau \in \{1, .., T\}$. This problem optimizes the objective function $\mathcal{F}_t(\cdot, S_t)$ at each time-step $t$, as specified in Eq. \eqref{eq:objective_P1}. This function depends on the active system $S_t$ and involves one of the following objectives: a) system response time in generating a plan, b) expected mission completion time while following the generated plan, and, c) energy efficiency of the generated plan.

Constraints in Eqs. \eqref{eq:P1_1} and \eqref{eq:P1_2} pertain to the UAV's dynamic constraints. Next, the constraint in Eq. \eqref{eq:P1_4} ensures that the agent's planned positional state $x_{t+\tau|t}^{\mathbf{p}}$ avoids residing inside the convex hull outlined by the firefront's path $C^i_{1:t}$, where $\triangle(.)$ denotes the convex-hull operator, and $C^i_{1:t}$ is the volume occupied by the firefront up to time-step $t$ as provided by the disaster warning system. 

Furthermore, Eq. \eqref{eq:P1_5} describes the decision dynamics $\Psi_t$ (i.e., a score on the importance of the engagement of each system at each point in time) monitored by the supervisory controller. This is influenced by the attention mechanism $A_t$, which depends on the agent and firefront states at time $t$. At each time-step $t$, the supervisory controller identifies the system in control $S_t$ that maximizes the decision-making score, as detailed in Eq. \eqref{eq:P1_6}. 
In the following section, we elaborate on how we addressed Problem (P1) by designing the various components of the proposed tripartite system.

\section{Cognitive Controller} \label{sec:approach}

\subsection{System 1 - Human-like Fast/Heuristic Planning} \label{ssec:s1}

As previously mentioned, System 1 encompasses the human process of creating a safe path that guides the UAV from its current position at time-step \( t \) to a designated goal region, all while circumventing firefronts. This involves the human operator assessing the current environmental state, such as the location and extent of firefronts, and subsequently inputting a sequence of waypoints to direct the UAV to its destination. This process, rooted in human intuition and causal reasoning, is rapid but may not yield an optimal trajectory in terms of mission completion time and energy efficiency. It is important to note that the outcome of this process is a reference path rather than a complete UAV trajectory. Consequently, the UAV must activate its low-level controller to track this reference path in accordance with its dynamical model. 
\begin{algorithm}
\caption{System 1  - Rapid UAV path planning}
\label{alg:alg1}
\begin{algorithmic}[1]
\State \textbf{Input:} UAV state $\hat{x}_{t|t}$, Goal region $G$, Firefronts $\hat{C}_t$
\State \textbf{Output:} Path from $\hat{x}_{t|t}$ to $G$ avoiding $\hat{C}_t$
\State Initialize Tree $Tr$ rooted at $\hat{x}_{t|t}$, $x_c \gets \hat{x}_{t|t}$
\While{$x_c \notin G$}
    \State $\tilde{x} \gets$ SampleRandomState($\mathcal{E}$)
    \State $x_{\text{nearest}} \gets$ NearestNeighbour($Tr$, $\tilde{x}$)
    \State $x_{\text{new}} \gets$ NewState($x_{\text{nearest}}, \tilde{x}$, StepSize)
    \If{PathFree($x_{\text{nearest}}, x_{\text{new}}, \hat{C}_t$)}
        \State Add $x_{\text{new}}$ to $Tr$
        	\State Add edge between $x_{\text{nearest}}$ and $x_{\text{new}}$
        	\State $x_c \gets x_{\text{new}}$
    \EndIf
    \If{Max Iterations reached}
        \State Break
    \EndIf
\EndWhile
\State \Return $\hat{x}_{t+\tau|t}$ by backtracking $T$
\end{algorithmic}
\end{algorithm}

To provide a systematic representation of the System 1 process without loss of generality, we have modelled it as a sampling-based path planning algorithm, akin to the Rapidly-exploring Random Trees (RRT) methodology \cite{Lavalle2001}. Nonetheless, alternative techniques possessing similar characteristics can also be employed to represent System 1, such as learning-based methods \cite{PapaioannouCDC2022}. The path-planning process carried out by System 1 is shown in Alg. \ref{alg:alg1}.

The algorithm takes as input the agent's state at the current time-step \( t \), the location of the goal region \( G \subset \mathcal{E} \) (i.e., a designated rectangular area), and the estimated occupied space \( \hat{C}_t \) of the firefronts up to time \( t \), i.e., \( \hat{C}_t = \bigcup_{i=1}^{N_t} C^i_{1:t} \). As shown, the algorithm operates by iteratively building a tree \( Tr \) from the agent's initial state \( \hat{x}_{t|t} \) towards the goal region \( G \). Each iteration involves randomly selecting a point in the space \( \tilde{x} \), and then extending the tree to a new point \( x_{\text{new}} \) from its nearest vertex \( x_{\text{nearest}} \) towards the sampled point, provided that this path is safe (i.e., the agent does not pass through the firefronts). This extension is governed by a predefined step size, ensuring that the tree gradually explores the space. The process repeats until the tree reaches the goal area or the maximum number of iterations is reached. This approach can be used to ensure an upper bound on the execution time, which is highly desirable in emergency response. Therefore, Alg. \ref{alg:alg1} is utilized with the primary objective of optimizing system response time.

In Alg.~\ref{alg:alg1}, the notation \( \hat{x}_{t+\tau|t} \) is used to signify that the generated result is a path and not a UAV trajectory. Subsequently, an LQR controller \cite{Yao2020} is utilized to track this resulting reference path $\hat{x}_{t+\tau|t}$ and provide the predictive UAV trajectory $x_{t+\tau|t}$ by minimizing the objective function \( J_\text{LQR} = \sum_{\tau} (x_{t+\tau|t}-\hat{x}_{t+\tau|t})^\top Q_\text{LQR} (x_{t+\tau|t}-\hat{x}_{t+\tau|t}) + \hat{u}_{t+\tau|t}^\top R_\text{LQR} \hat{u}_{t+\tau|t} \), where \( \hat{u}_{t+\tau|t} = -K_\text{LQR}(x_{t+\tau|t}-\hat{x}_{t+\tau|t}) \), the matrix \( K_\text{LQR} \) is the LQR gain, the weight matrix \( Q_\text{LQR} \) penalizes deviations from the reference trajectory, and \( R_\text{LQR} \) is a weight matrix that penalizes the use of control inputs.

\subsection{System 2 - Slow/Deliberative Optimization}  \label{ssec:s2}

System 2 is a fully autonomous system that utilizes Model Predictive Control (MPC) to generate an optimal and safe UAV trajectory. It optimizes a specific objective function, which in this work is either the mission completion time (i.e., the trajectory that guides the UAV to the goal region in the least amount of time) or the energy efficiency (i.e., the trajectory that minimizes energy usage by reducing abrupt changes between consecutive control inputs). Therefore the objective function is formulated as $\mathcal{J}(X_T,U_T)=$
\begin{equation} \label{eq:S2_mission_objective}
     \kappa_1 \sum_{\tau=1}^{T}\| x^{\mathbf{p}}_{t+\tau|t}-G_o\|^2_2 
    + \kappa_2 \sum_{\tau=1}^{T} \|u_{t+\tau|t}-u_{t+\tau-1|t}\|^2_2,
\end{equation}

\noindent where $G_o \in \mathbb{R}^3$ is the centroid of the goal region $G$, and $\kappa_i > 0, i\in \{1,2\}$ are a design parameters which controls the preference between the two objectives.
\begin{algorithm}
\vspace{-3mm}
\begin{subequations}
\begin{align} 
&\hspace*{-3mm}\textbf{(P2)}~\texttt{System 2 - Optimal Planning} & \notag\\
& \hspace*{-3mm}~~~~\underset{\{u_{t+\tau|t},x_{t+\tau|t}\}_{\tau=1}^{T}}{\min} ~\mathcal{J}(X_T,U_T), &  \hspace*{-10mm} \label{eq:objective_P2} \\
&\hspace*{-3mm}\textbf{subject to: $\tau \in [1,..,T]$} ~  &\nonumber\\
&\hspace*{-3mm} x_{t+\tau|t} = \Phi x_{t+\tau-1|t} + \Gamma u_{t+\tau|t} & \hspace*{-10mm} \label{eq:P2_1}\\
&\hspace*{-3mm} x_{t|t} = x_{t|t-1} & \hspace*{-10mm} \label{eq:P2_2}\\
&\hspace*{-3mm} \text{dot}(\alpha_{i,l},x^{\mathbf{p}}_{t+\tau|t}) \geq \beta_{i,l} + \epsilon - M b_{\tau,i,l} &\hspace*{1mm} \forall \tau,i, l \label{eq:P2_3}\\
&\hspace*{-3mm} \sum_{l=1}^6 b_{\tau,i,l} \le 5 &\hspace*{1mm} \forall \tau,i \label{eq:P2_4}\\
&\hspace*{-3mm} b_{\tau,i,l} \in \{0,1\}, ~i \in \{1,..,\hat{N}_t\} &\hspace*{1mm} \forall \tau,i,l \label{eq:P2_5}\\
&\hspace*{-3mm} x_{t+\tau|t} \in \mathcal{X}, ~u_{t+\tau|t} \in \mathcal{U} &\hspace*{1mm} \forall \tau \label{eq:P2_6}
\end{align}
\end{subequations}
\vspace{-5mm}
\end{algorithm}

Problem (P2) is essentially a rolling-horizon MPC problem formulated as a mixed integer quadratic program (MIQP) consisting of linear and binary constraints, which can be solved to optimality using off-the-shelf optimization tools. However, its computational complexity increases with the number of constraints which in this case depend on the number of firefronts that need to be avoided and the length $T$ of the planning horizon. Suppose, that the number of active firefronts inside the disaster environment up to time-step $t$ is given by $\hat{N}_t$ which is essentially the size of the set $\hat{C}_t$ defined in Sec. \ref{ssec:s1}. Given the cuboid representation $C_i \in \hat{C}_t$ of firefront $i \in \{1,..,\hat{N}_t\}$, which is associated with the parameter set  $\{\alpha_{i,l},\beta_{i,l}\}_{l=1}^6$, the constraints in Eqs. \eqref{eq:P2_3} - \eqref{eq:P2_4} formulate firefront avoidance constraints utilizing the binary variable $b_{\tau,i,l}$, as shown in our previous works \cite{PapaioannouTMC2023,PapaioannouCDC2023}. Specifically, to generate a safe predictive trajectory which avoids all firefronts $i$ we require that $ x^{\mathbf{p}}_{t+\tau|t} \notin \mathcal{C}_i,~\forall i, \forall \tau$. This condition, is satisfied at time-step $\tau$ when:
\begin{equation}\label{eq:collision}
  \exists l \in \{1,..,6\}: \text{dot}(\alpha_{i,l},x^{\mathbf{p}}_{t+\tau|t}) > \beta_{i,l}
 \end{equation}

\noindent To accomplish this for all time-steps and all firefronts the constraint in Eq. \eqref{eq:P2_3} uses the binary variable $b_{\tau, i, l}$ to check if Eq. \eqref{eq:collision} is violated, where $M$ is a big positive constant, and tolerance $\epsilon>0$. Constraint \eqref{eq:P2_4} counts the number of violations and ensures that is less or equal to $6$, thereby enforcing Eq. \eqref{eq:collision}. 


\subsection{Supervisory Controller - System 1 and 2 Orchestration}  \label{ssec:s3}
The supervisory controller decides which system to engage by assessing their performance on a number $n$ of predefined attributes $a \in \{1,..,n\}$. In this work, we focus on the characterisation of the two systems on $n=3$ attributes including $a=1$, representing the system response time (i.e., the time required for trajectory generation), $a=2$, denoting the mission completion time (i.e., the time taken to reach the goal region given the agent follows the generated plan), and $a=3$, indicating the energy usage required by the UAV to execute the planned trajectory (i.e., the sum of squared deviations between consecutive control inputs). 

The performance of each system (System 1 and System 2) across the $n$ specified attributes is represented by the 2 by 3 matrix $Q$. This matrix facilitates a comprehensive comparison of the two systems (represented by rows) with respect to the aforementioned attributes (represented by columns), where $Q(S,a)$ quantifies the performance of system $S \in \{1, 2\}$ in relation to attribute $a$. For example, based on the system characteristics outlined in Sec. \ref{ssec:s1} and Sec. \ref{ssec:s2}, the performance matrix $Q$ can be defined as $Q = \left[ \begin{smallmatrix} 0.8 & 0.5 & 0.3\\ 0.4 & 0.9 & 0.9 \end{smallmatrix} \right]$. This indicates that System 1 performs better in terms of response time, while System 2 excels in mission completion time and energy efficiency.
Inspired by the foundational principles of human behavioral decision theories \cite{Slovic1977}, we develop a probabilistic attention mechanism. This mechanism, based on the state of the environment and the agent, dynamically modulates the agent's focus towards the most relevant attribute necessary for completing the mission. Subsequently, this process aims to optimize the engagement of the most appropriate system (System 1 or System 2) for the task at hand. 
Specifically, the attention probability $p_a, a\in \{1,..,n\}$ given to each attribute $a \in \{1,..,n\}$ follows a Dirichlet probability density function with (attention) parameters $\Xi=(\xi_a| a \in \{1,..,n\})$ given by:
\begin{equation} \label{eq:dir}
	f_D(P|\Xi) = \frac{1}{\mathcal{B}(\Xi)}\prod_{a} p_a^{\xi_a-1},
\end{equation}
\noindent where $P = [p_1,p_2,p_3]$,  the normalising constant $\mathcal{B}(.)$ is the multivariate beta function, $\xi_a > 0, \forall a$, $p_a \in [0,1], \forall a$, and finally $\sum_{a} p_a = 1$. Accordingly, we define the attention random vector $A$ as a column vector of size $n$, having the property that exactly one element has the value 1 and the others have the value 0. The particular element having the value 1 at some index $a$ indicates that attribute $a$ has been selected with probability $p_a$, and therefore $A$ distributed according to:
\begin{equation}\label{eq:categorical}
	f_A(A|P) = \prod_a p_a^{A(a)},
\end{equation}
\noindent where $A(a) \in \{0,1\}$, and $\sum_a A(a) = 1$. Subsequently, we can draw attention to a particular attribute by sampling the probability distribution in Eq. \eqref{eq:categorical}.

 The agent's decision-making behavior during mission execution is governed by designing or learning the dynamical behavior of the attention parameters $\Xi$. In this study, the attention parameters are time-variant ($\Xi_t$), leading to a dynamic allocation of attention probabilities to each of the three key attributes, under the following assumptions:

 \begin{figure}
	\centering
	\includegraphics[width=\columnwidth]{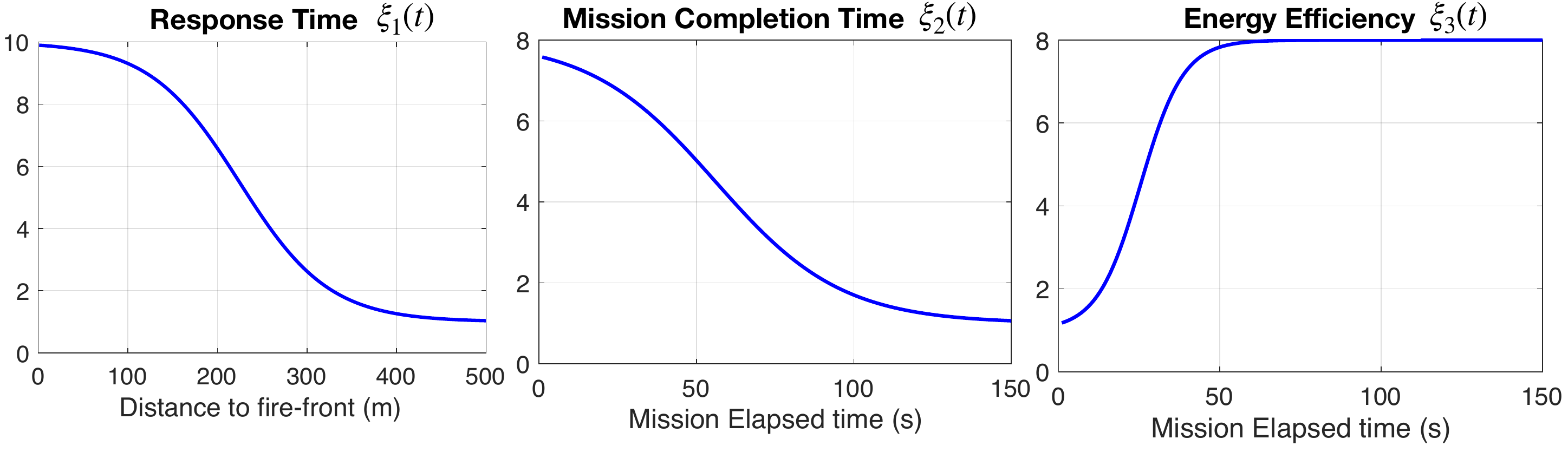}
	\caption{The figure illustrates the dynamic behavior of the attention parameters $\xi_1(t), ~\xi_2(t),$ and $\xi_3(t)$ based on the following configuration  $w_{1a}=40$, $w_{1b}=0.02$, $w_{1c}=9$, $w_{1d}=1$, $w_{2a}=10$, $w_{2b}=0.05$, $w_{2c}=7$, $w_{2d}=1$, $w_{3a}=10$, $w_{3b}=0.15$, $w_{3c}=7$, and $w_{3d}=1$.}	
	\label{fig:fig3}
	\vspace{-5mm}
\end{figure}

\noindent \textbf{System Response Time} - $\xi_1(t)$:
 The prioritization of system response time escalates in correlation with the proximity of the UAV to emergent firefronts. In scenarios where a firefront genesis event occurs in close vicinity to the UAV, an expedited response becomes critical. This necessitates the rapid generation of UAV trajectories, prioritizing immediacy over optimality, to mitigate the risks posed by the unpredictable progression of the firefront.
 
\noindent \textbf{Mission Completion Time} -  $\xi_2(t)$:
In the initial phase of the mission, it is crucial to determine a trajectory that enables the UAV to reach the target area in the shortest possible time. This approach is essential in maximizing the probability of locating and rescuing survivors. However, as the mission progresses, the emphasis placed on minimizing mission completion time is progressively reduced.

\noindent \textbf{Energy Efficiency} - $\xi_3(t)$:
 Over time, the UAV's battery health deteriorates owing to irreversible physical and chemical degradation, leading to decreased stability. Consequently, after a predefined duration of operation, the flight controller transitions to an energy conservation mode. This shift prompts a gradual elevation in the importance assigned to energy efficiency in the decision-making process.

\noindent The strategy outlined above for the dynamic allocation of the attention parameters $\Xi_t = \left( \xi_1(t),\xi_2(t),\xi_3(t) \right)$ for the 3 attributes is implemented in this work as follows:
\begin{subequations}\label{eq:KSI}
\vspace{+1.5mm}
\begin{align}
   &\xi_1(t) = w_{1c} -  \left[ \frac{w_{1c}}{1 + w_{1a} \exp \left(-w_{1b}  (d_t - w_{1a})\right)} \right] + w_{1d},  \\
   & \xi_2(t)= w_{2c} - \left[ \frac{w_{2c}}{1 + w_{2a} \exp(-w_{2b} (t - w_{2a}))} \right] + w_{2d}, \\
    & \xi_3(t) = w_{3c} \left[ \frac{1}{1 + w_{3a} \exp(-w_{3b}  (t - w_{3a}))} \right] + w_{3d}. 
\end{align}
\vspace{+1.5mm}
\end{subequations}

The tuning hyperparameters $w_{ic}$ and $w_{id}$, for each $i \in \{1, .., 3\}$, are set to establish the upper and lower limits of the attention parameters, respectively. In contrast, $w_{ia}$ and $w_{ib}$ adjust the attention response in relation to the input stimuli, and finally, the term $d_t = ||x_t^{\mathbf{p}} - \hat{y}_t^{\mathbf{p}}||_2$ quantifies the Euclidean distance at time-step $t$ between the UAV's current position, represented as $x_t^{\mathbf{p}}$, and the location of the nearest firefront, $\hat{y}_t^{\mathbf{p}}$, which emerged at time-step $t$. The dynamic behavior of $\Xi_t$ based on Eq. \eqref{eq:KSI} for a specific set of hyperparameters is illustrated in Fig. \ref{fig:fig3}.

\begin{figure*}
	\centering
	\includegraphics[width=\textwidth]{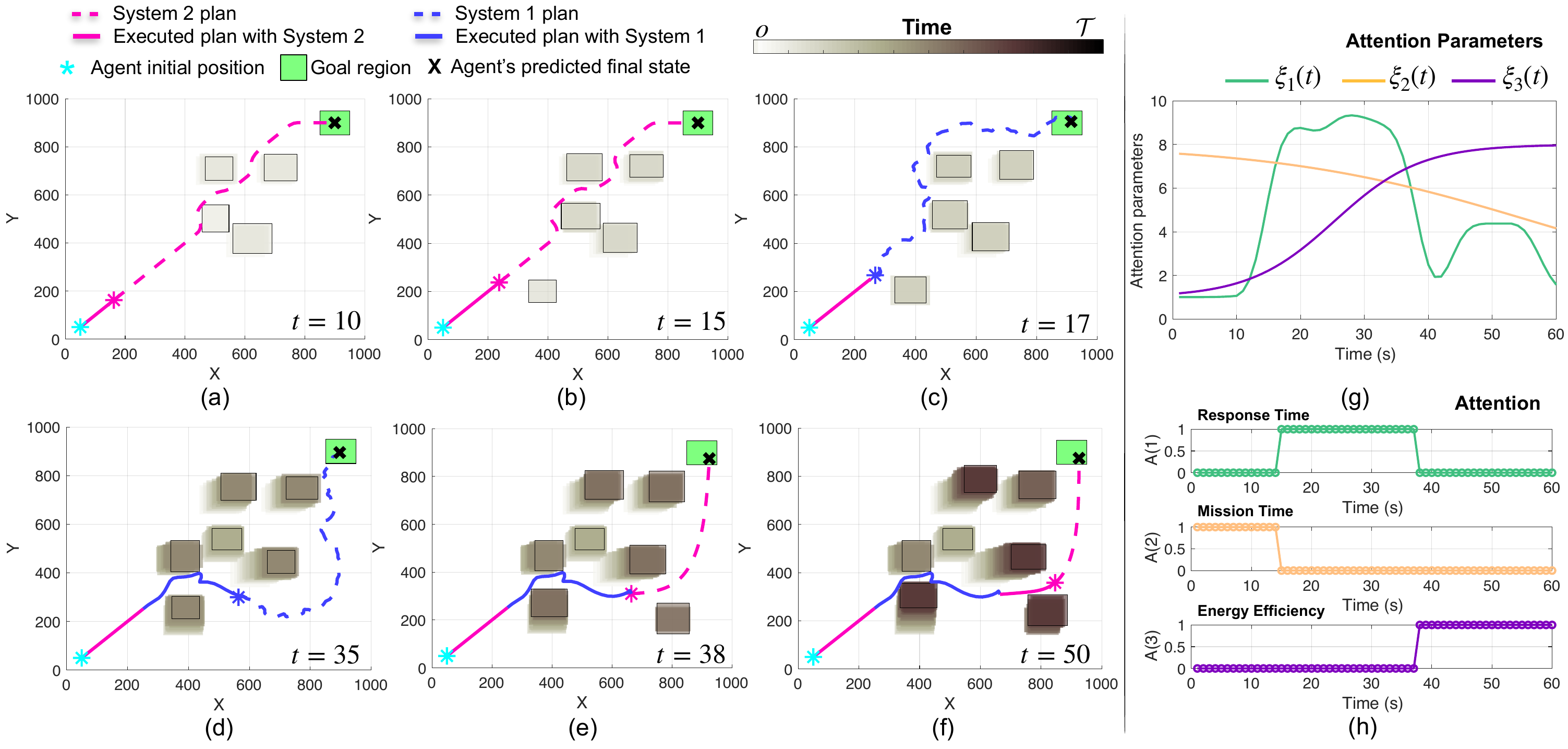}
	\caption{The figure illustrates the proposed approach in a simulated disaster scenario, demonstrating the operation of the proposed tripartite cognitive architecture. (a)-(f) Trajectory planning using a combination of System 1 and System 2 based on attribute performance, (g) The evolution of attention parameters during the mission, (h) The activation of different attributes based on the attention vector $A_t$.}
	\label{fig:fig_r1}
	\vspace{-5mm}
\end{figure*}

The decision dynamics $\Psi_t$ are described by the following state-space model:
\begin{equation}\label{eq:Psi}
	\Psi_t = H \Psi_{t-1} + BQA_t,
\end{equation}
\noindent where $H$ is the state transition matrix, modeling the decision-making memory effects and hence the dependence of the current decision on previous ones. The matrix $B = \left[ \begin{smallmatrix} 1 & -1 \\ -1 & 1 \end{smallmatrix} \right]$ is a constant matrix. When multiplied with the system performance matrix $Q$, it quantifies the differential disparity between each system in relation to a specific attribute. Finally, $A_t$ represents the attention process, distributed according to $A_t \sim f_A(.|P_t)$, with $P_t \sim f_D(.|\Xi_t)$, where $\Xi_t$ evolves in accordance with Eq. \eqref{eq:KSI}. 
Consequently, the supervisory controller, at each time-step $t$, determines the system in control by maximizing the decision-making score as:
\begin{equation}\label{eq:maxPsi}
	S_t = \underset{i \in \{1,2\}}{\arg\max} ~\Psi_t(i).
\end{equation}

\section{Evaluation}\label{sec:Evaluation}

\subsection{Simulation Setup} \label{ssec:sim_setup}
We evaluate the proposed approach by simulating the disaster scenario described in Sec. \ref{sec:problem} according to the setup described below. For illustrative purposes, and in order to simplify the analysis we demonstrate our results in a top-down view (i.e., 2-dimensional settings). Therefore, the disaster environment $\mathcal{E}$ has dimensions $1$ km $\times$ $1$ km, and the UAV agent evolves (assuming a fixed altitude) according to the dynamics described in Eq. \eqref{eq:kinematics}, with parameters $\Delta T$, $\zeta$, and $m$ are set to 1s, 0.2, and 1.05kg respectively. The UAV agent can reach a maximum velocity of 15m/s, utilizing an input control force in the range of $[-7.5,7.5]$N. The UAV's initial position is set to $(x,y) = (50, 50)$, and the goal region $G$ is represented as a rectangular region centered at $G_o = (900, 900)$ with size 100m $\times$ 100m.
The firefront process dynamics are given by $p_b = 0.1$, $p_s = 0.75$ and the birth density $b_t(.)$ is uniform inside the environment. In addition, $\bar{y}^{\nu}=3$m/s, $\bar{y}^{d}=\pi/4$, $\sigma_\nu=0.8$, $\sigma_d = \pi/20$, and finally $\gamma=0.8$. Moreover, the extend (i.e., size) of each firefront at each time-step is represented as a rectangle with dimensions uniformly sampled from the range $[100,250]$m. Subsequently, System 1 is configured for running for maximum of 500 iterations, with a step-size of 15m, and tracked with an LQR controller with $Q_\text{LQR}=2I_{4 \times 4}$, and $R_\text{LQR}=0.1I_{2 \times 2}$ Then the parameters in System 2's objective function in Eq. \eqref{eq:S2_mission_objective} are set to $(\kappa_1 = 1,\kappa_2=0.001)$ if the attention is on the mission completion time attribute, and conversely set to $(\kappa_1 = 0.001,\kappa_2=1)$ if the attention is on the energy efficiency. Then, the parameter $M$ in Problem (P2) is set to $M=10^5$, and $T=80$. Regarding the supervisory controller, the performance matrix is set to $Q = \left[ \begin{smallmatrix} 0.8 & 0.5 & 0.3\\ 0.4 & 0.9 & 0.9 \end{smallmatrix} \right]$, and the attention parameters $\Xi_t = \left( \xi_1(t),\xi_2(t),\xi_3(t) \right)$ take their values according to Eq. \eqref{eq:KSI} with parameters $w_{1a}=40$, $w_{1b}=0.02$, $w_{1c}=9$, $w_{1d}=1$, $w_{2a}=10$, $w_{2b}=0.05$, $w_{2c}=7$, $w_{2d}=1$, $w_{3a}=10$, $w_{3b}=0.15$, $w_{3c}=7$, and $w_{3d}=1$, which generate the profiles illustrated in Fig. \ref{fig:fig3}. Finally, the state transition matrix $H$ of the decision-making dynamics shown in Eq. \eqref{eq:Psi} is set to $H = 0.5I_{2 \times 2}$. Finally, Problem (P2) is solved using the Gurobi solver, running on a 3.2GHz CPU system.

\subsection{Results}
Figure \ref{fig:fig_r1} illustrates the proposed approach in a simulated disaster scenario, unfolding over $\mathcal{T} = 80$ time-steps. In this figure, the UAV's executed trajectory up to time-step $t$ is depicted as a solid line (with blue and pink colors representing System 1 and System 2, respectively), while the generated plan at time $t$ is shown as a dotted line. The UAV's state at the current time-step $t$ is indicated by the $\ast$ symbol, and its final predicted state is marked with $\times$. The firefronts are represented by the shaded rectangular areas.

Specifically, Fig. \ref{fig:fig_r1}(a) displays the UAV's executed and predicted trajectory at time-step $t=10$. The generated plan at time-step $t=10$ illustrates the trajectory that leads the UAV to the goal region in the shortest possible time (analogous to the shortest path to the destination). The activation of System 2 up to this point is guided by the evolution of the attention parameters $\Xi_t = (\xi_1(t), \xi_2(t), \xi_3(t))$, as depicted in Fig. \ref{fig:fig_r1}(g). These parameters, in turn, influence the attention probabilities and the decision-making dynamics.

\begin{figure}
	\centering
	\includegraphics[width=\columnwidth]{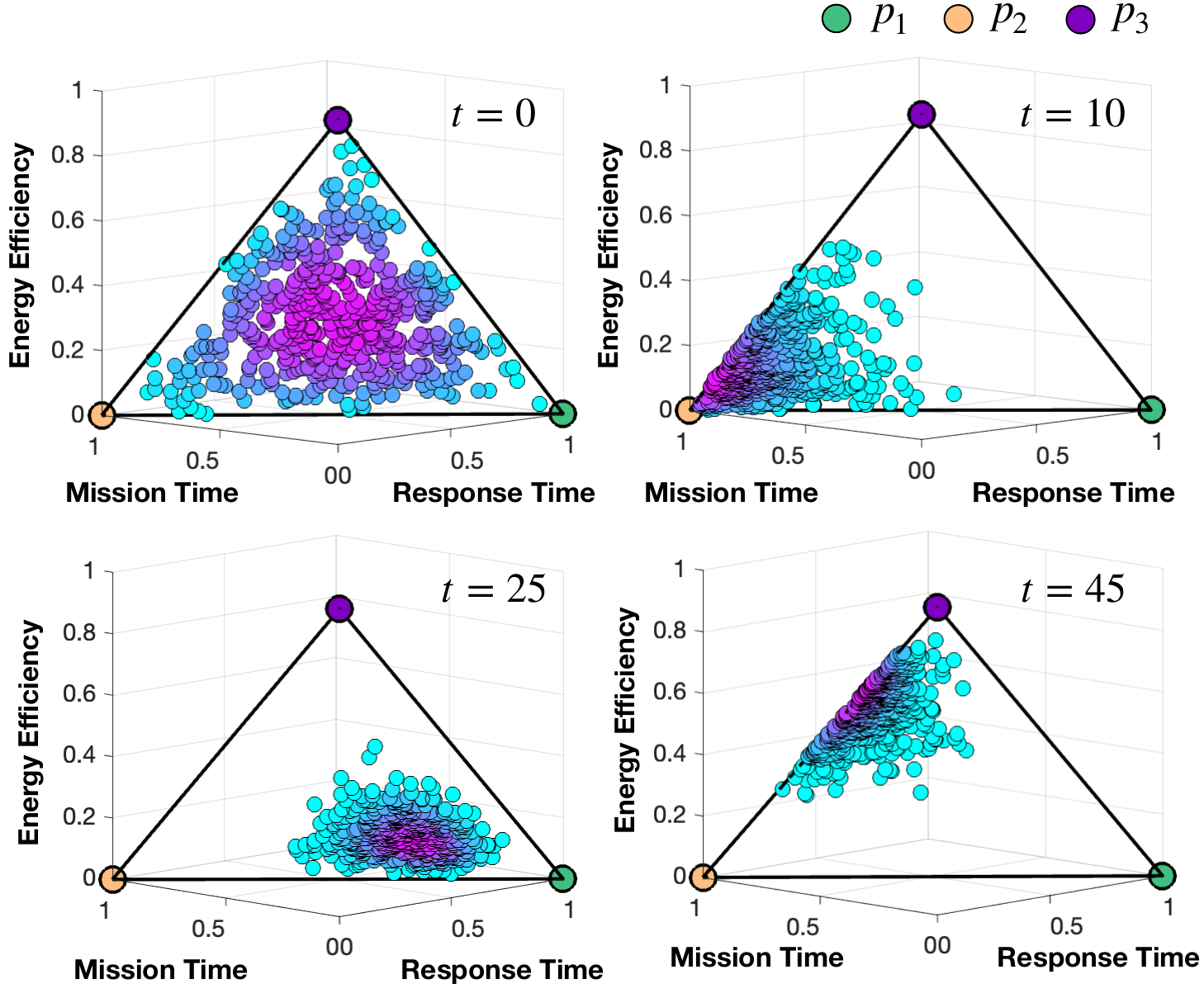}
	\caption{The figure illustrates the probability distribution on the attention probabilities $(p_1,p_2,p_3)$ i.e., Eq. \eqref{eq:dir} assigned to the 3 attributes at different points in time during the mission.}	
	\label{fig:fig_r2}
	\vspace{-5mm}
\end{figure}

\begin{figure}
	\centering
	\includegraphics[width=\columnwidth]{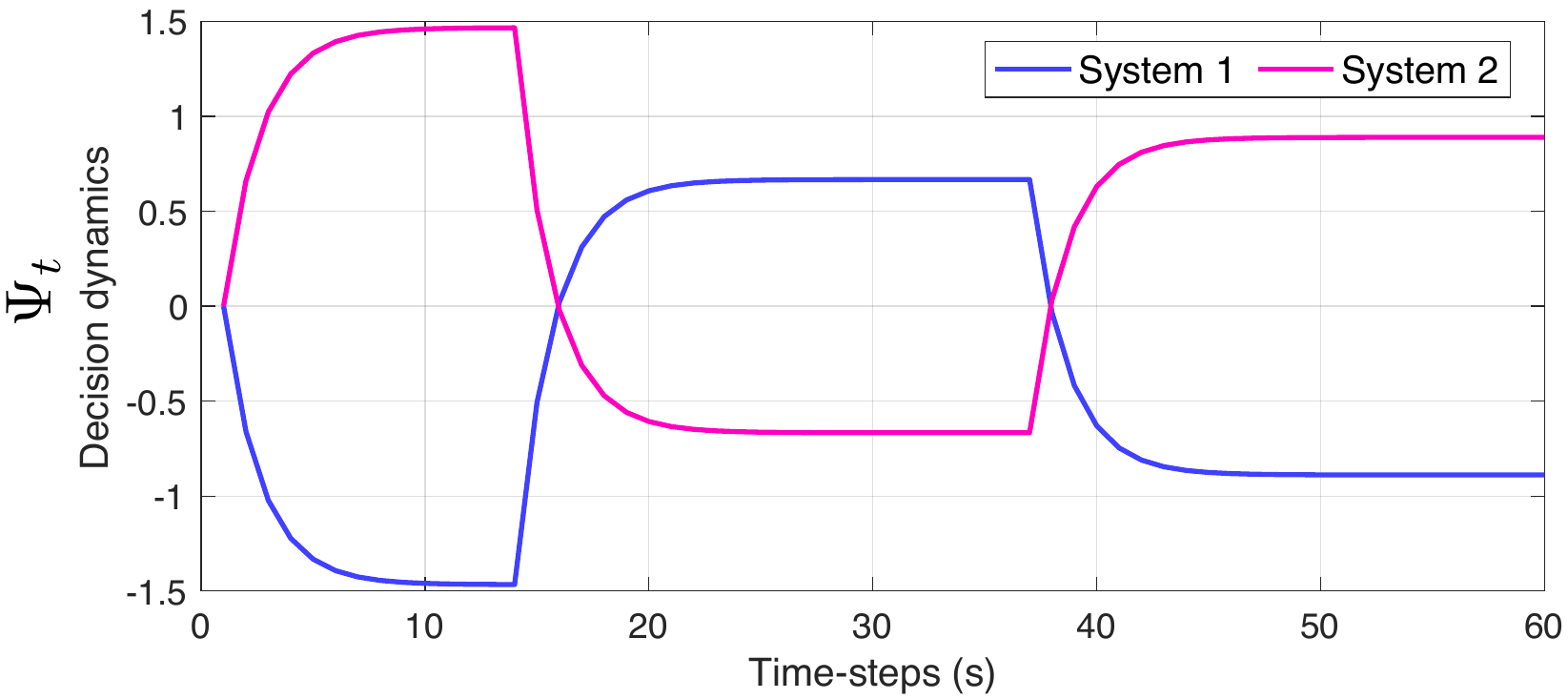}
	\caption{The figure shows the decision-making dynamics $\Psi_t$. At each time-step the supervisory controller decides which system to engage by maximising the decision-making score.}	
	\label{fig:fig_r3}
	\vspace{-3mm}
\end{figure}

To illustrate this, note that the value of the attention parameter $\xi_2(t)$, corresponding to the mission completion time attribute, takes the highest value at $t=10$ amongst the attention parameters, as shown in Fig. \ref{fig:fig_r1}(g). This, in turn, shifts the probability distribution of the three attributes in favour of mission completion time. Fig. \ref{fig:fig_r2} demonstrates this for $t=10$, displaying the probability distribution, i.e., Eq. \eqref{eq:dir}, of the attention probabilities $P = [p_{1}, p_{2}, p_{3}]$ at this time-step. Specifically, the figure presents 200 samples of $P_t \sim f_D(.|\Xi_t)$, highlighting a concentration around $p_2$. This concentration directs the attention focus towards the second attribute via Eq. \eqref{eq:categorical}, i.e., the operation $A_t \sim f_A(.|P_t)$ samples an attention vector, which with high probability has its second element activated, i.e., $A = [0,1,0]^\top$, as depicted in Fig. \ref{fig:fig_r1}(h) for this time-step. The attention is then integrated with the preference matrix $Q$ in the decision-making dynamics of Eq. \eqref{eq:Psi}, illustrated in Fig. \ref{fig:fig_r3}. The supervisory controller uses this to determine the controlling system according to Eq. \eqref{eq:maxPsi}, in this case, System 2, as shown. It is important to observe, from Fig. \ref{fig:fig_r2}, that initially ($t=0$), the distribution of attention probabilities across the three attributes is uniform but shifts towards specific attributes over time to optimize the system's overall performance.

Subsequently, the agent operates using System 2 until a new firefront appears in close vicinity at $t=15$, as indicated in Fig. \ref{fig:fig_r1}(b). From this point onwards, the attention shifts towards fast response attributes to mitigate the risk to the UAV due to its proximity to newly spawned firefronts. This shift in focus is evidenced by the increasing magnitude of the $\xi_1(t)$ parameter, as shown in Fig. \ref{fig:fig_r1}(g), and the corresponding attention shift depicted in Fig. \ref{fig:fig_r1}(h). Consequently, at time-step $t=17$, control is transferred to System 1 for quick action and rapid trajectory generation, as shown in Fig. \ref{fig:fig_r3} (i.e., $\arg\max_i \Psi_t(i) = 1$ at $t=17$). Figure \ref{fig:fig_r1}(c) illustrates the trajectory generated by System 1, marked with a dotted blue line. Note that this trajectory is not the most optimal in terms of mission completion time. The UAV continues operating under System 1 until $t=37$, as indicated by the attention parameters in Figs. \ref{fig:fig1}(g)(h) and the decision dynamics in Fig. \ref{fig:fig_r3}. It switches back to System 2 at $t=38$ to optimize its energy efficiency, as illustrated in Fig. \ref{fig:fig_r1}(e)(f) and Fig. \ref{fig:fig_r3}.

Finally, we compare the performance of the proposed tripartite system against System 1 and System 2, aiming to quantify the performance improvements relative to scenarios where the mission is executed exclusively with either the fast and heuristic System 1 or the slow and deliberative System 2. To achieve this, we conducted a Monte-Carlo simulation by generating 200 random mission configurations. These configurations simulate disaster environments, as described in Sec. \ref{ssec:sim_setup}, and include variations in the UAV's initial position and the goal region. We then executed each mission using a UAV agent guided by a) the proposed tripartite system, b) only System 1, and c) only System 2. Our objective is to characterize the effectiveness of the proposed technique in terms of its performance on response and mission time, as well as energy efficiency.
The response time is assessed based on the agent's ability to successfully complete the mission and reach the goal region. A mission is considered failed if the agent is trapped by a firefront, i.e., the distance between the agent and a newly emerged firefront becomes less than a predefined safety threshold (denoted as $d_\text{safe}=20$m). This may occur if the trajectory generation computation is not sufficiently rapid.

\begin{figure}
	\centering
	\includegraphics[width=\columnwidth]{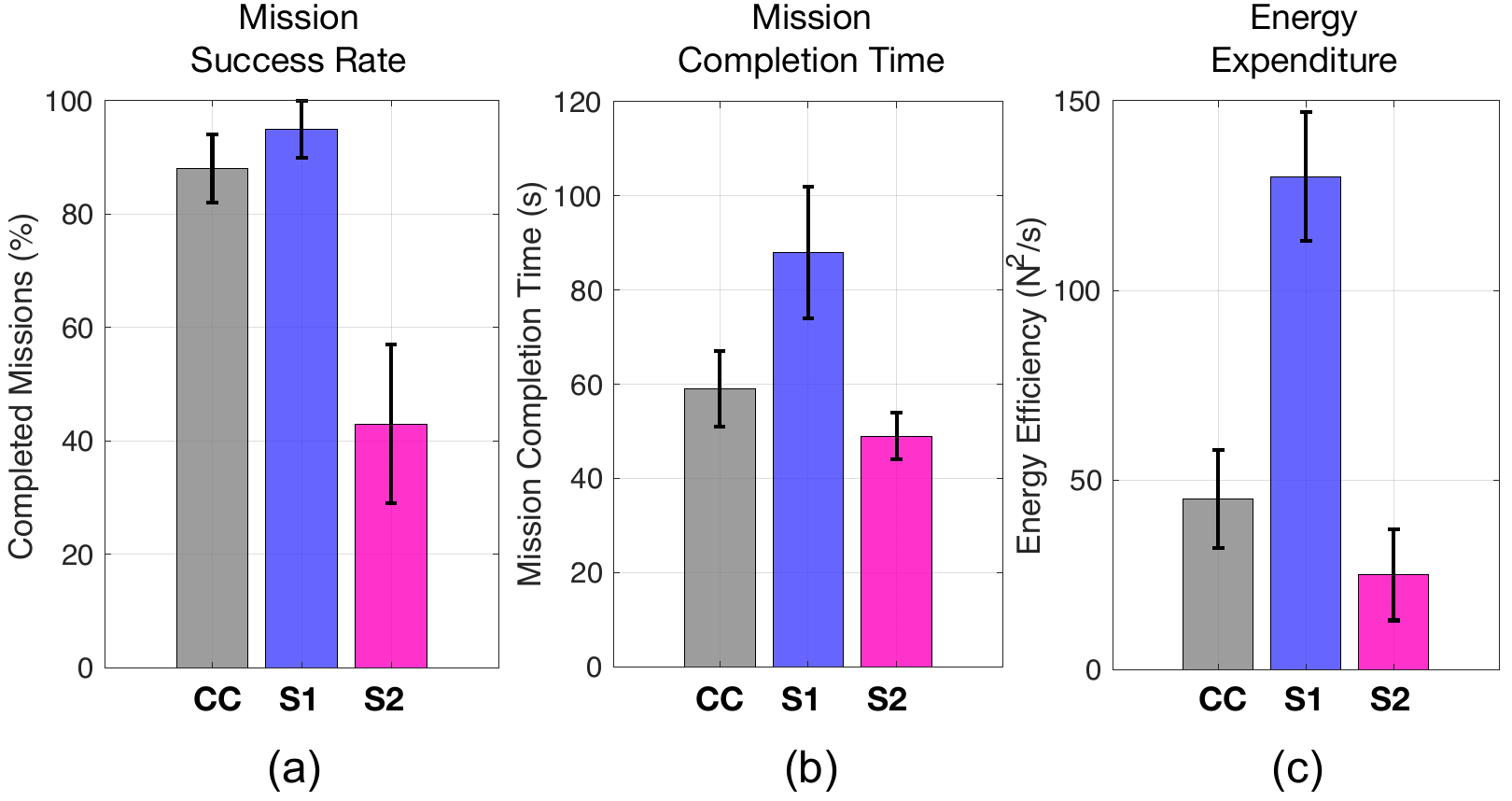}
	\caption{Comparison between the proposed tripartite Cognitive Controller (CC), and an agent that only utilized either System 1 (S1) or System 2 (S2).}	
	\label{fig:fig_r5}
	\vspace{-3mm}
\end{figure}

To test this hypothesis, we set the firefront propagation time to $T_f=0.5$s, based on the average trajectory generation times of System 1 (approximately 155ms) and System 2 (approximately 5s). This setting makes the response time of System 1 three times faster compared to the firefront evolution, whereas System 2 is ten times slower. At each time-step, we measure the time $T_s$ required for each system to generate a trajectory. Before transitioning the agent to the next state, we propagate the firefronts $\lceil \frac{T_s}{T_f} \rceil$ times in sequence and check whether the distance between the propagated firefronts and the agent's idle position (i.e., the agent remains still while computing its plan) has dropped below $d_\text{safe}$. If this is the case, we mark the mission as a failure. 

Fig. \ref{fig:fig_r5}(a) reveals that a UAV agent managed to complete approximately $40\%$ of the missions using System 2 (shown as S2), compared to around $95\%$ with System 1 (shown as S1). The proposed technique, shown as CC (i.e. Cognitive Control), achieved a similar mission completion rate to System 1, around $90\%$, as shown. This discrepancy is due to the attention response to close-proximity firefronts and the reliance on previous decisions, which can be fine-tuned for faster switching times. Fig. \ref{fig:fig_r5}(b) presents the average mission completion time. In this metric, System 2 outperforms System 1. The proposed approach shows comparable results in this domain, while also achieving a higher mission success rate, as previously discussed in Fig. \ref{fig:fig_r5}(a). The discrepancy between the proposed approach and System 2 arises from the number of times our methodology switched to the heuristic System 1 in response to emergent firefronts. Finally, Fig. \ref{fig:fig_r5}(c) displays similar trends in energy efficiency.

\section{Conclusion} \label{sec:conclusion}
Drawing on the principles of dual process theory (DPT), we propose a cognitive control architecture that combines human-like responses with analytical machine intelligence. We formulate the problem of decision-making  and planning as a hierarchical optimization problem within a DPT-inspired framework. Moreover, an attention-based supervisory controller, informed by the agent's state and environmental inputs, directs control to the optimal system for planning. This integrated approach is particularly effective in handling complex tasks, enabling rapid adaptation in challenging situations. 
\balance
\bibliographystyle{IEEEtran}
\bibliography{IEEEabrv,wcci2024}

\end{document}